\documentclass[11pt,a4paper]{article}
\usepackage[hyperref]{acl2019}
\usepackage{times}
\usepackage{latexsym}
\usepackage{amsmath,amsfonts,amssymb}
\usepackage{url}
\usepackage{graphicx}
\usepackage[multiple]{footmisc}
\usepackage[T1]{fontenc}
\usepackage[utf8]{inputenc}
\usepackage[italian]{babel}

\aclfinalcopy 

\newcommand\cls{\texttt{[CLS]}}

\newcommand\bertintent{\texttt{Bert-Intent}}
\newcommand\bertslot{\texttt{Bert-Slot}}
\newcommand\bertjoint{\texttt{Bert-Joint}}
\newcommand\snipsita{\texttt{SLU-IT}}

\title{Multi-lingual Intent Detection and Slot Filling in a Joint BERT-based Model}

\author{Giuseppe Castellucci \, \, Valentina Bellomaria \, \, Andrea Favalli \, \, Raniero Romagnoli\\
  Language Technology Lab \\
  Almawave \\
  Via di Casal Boccone 188, 00137 Rome, Italy \\
  \texttt{\{last\_name\}@almawave.it}
}

\date{}

\begin{document}
\maketitle
\begin{abstract}
Intent Detection and Slot Filling are two pillar tasks in Spoken Natural Language Understanding. Common approaches adopt joint Deep Learning architectures in attention-based recurrent frameworks.
In this work, we aim at exploiting the success of ``recurrence-less'' models for these tasks. 
We introduce \bertjoint{}, i.e., a multi-lingual joint text classification and sequence labeling framework. The experimental evaluation over two well-known English benchmarks demonstrates the strong performances that can be obtained with this model, even when few annotated data is available. Moreover, we annotated a new dataset for the Italian language, and we observed similar performances without the need for changing the model.
\end{abstract}

\section{Introduction}
\label{sec:intro}
Recently, conversational interfaces, e.g., Google's Home or Amazon's Alexa, are becoming pervasive in daily life. As an important part of any conversation, language understanding aims at extracting the meaning a partner is trying to convey.
Spoken Language Understanding (SLU) plays a critical role in such a scenario. Generally speaking, in SLU a spoken utterance is first transcribed, then semantics information is extracted.

\begin{figure}[t]
  \centering
    \includegraphics[width=0.46\textwidth]{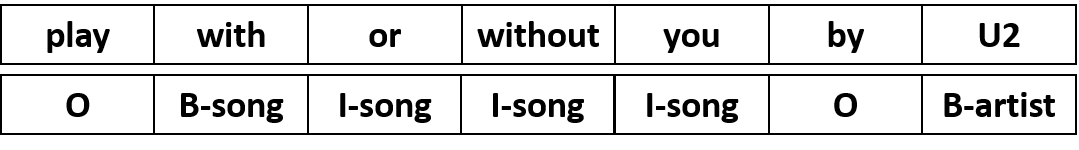}
  \caption{An example of Slot Filling in IOB format for a sentence with intent \textit{PlayMusic}.}
  \label{fig:slu-example}
\end{figure}

In this work, we concentrate on language understanding, i.e., extracting a semantic ``frame'' from a transcribed user utterance. Typically, this involves two tasks: Intent Detection (ID) and Slot Filling (SF) \cite{tur2011}. The former tries to classify a user utterance into an intent, i.e., the purpose of the user. The latter tries to find what are the ``arguments'' of such intent. As an example, let us consider Figure \ref{fig:slu-example}, where the user asks for playing a song (\texttt{Intent=PlayMuysic}) (\textit{with or without you}, \texttt{Slot=song}) of an artist (\textit{U2}, \texttt{Slot=artist}).

Common approaches address the ID and SF tasks in joint Deep Learning architectures (e.g., \cite{liu2016,slotgated2018}). In particular, encoder-decoder models \cite{seq2seq2014} and/or recurrent neural networks (RNN) with attention \cite{bahdanau2014} are trained on the task of predicting at the same time intents and slots.
Recently, recurrence-less models \cite{vaswani2017} shifted the attention on a neural-based computation for natural language which is not based on the typical recurrent processing happening in RNNs. Based on this idea, BERT \cite{devlin2018} models multiple tasks with a unique deep attention-based architecture.

In this work, we extend BERT by jointly modeling the ID and SF tasks. In particular, we define a joint text classification and sequence labeling framework based on BERT, i.e., \bertjoint{}. Specifically, we base our model on the BERT pre-trained representations and we add on top of it a text classifier and a sequence labeler, which are trained jointly over a unique loss function. The experimental evaluation shows that the proposed approach can achieve strong performances on the well-known ATIS \cite{atis1990} dataset. Moreover, it can reach the state-of-the-art on the newer SNIPS \cite{snips2018} dataset. Finally, we annotated a new dataset for the ID and SF tasks in Italian. We will show the applicability of \bertjoint{} also over this dataset without the need for adapting the model.

Following, in section \ref{sec:approach} the proposed approach will be presented, while in section \ref{sec:experiments} the experimental evaluation will be provided. In section \ref{sec:related} related works will be discussed. Finally, section \ref{sec:conclusion} will derive the conclusions.

\section{Joint Modeling of Intents and Slots within the BERT Framework}
\label{sec:approach}

\begin{figure}[t]
  \centering
    \includegraphics[width=0.18\textwidth]{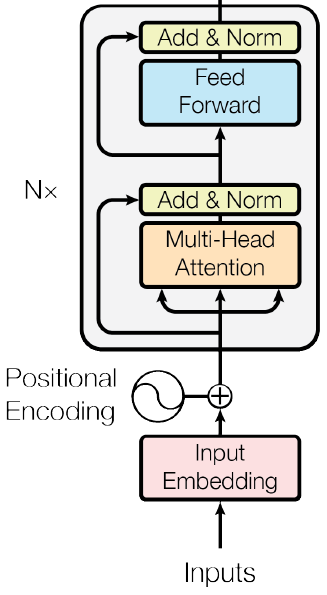}
  \caption{The Transformer encoder structure.}
  \label{fig:transformer}
\end{figure}

In this section, we present \bertjoint.
First, BERT is briefly introduced in section \ref{sec:bert}. In section \ref{sec:jbert} the proposed joint model is described.

Let us consider we have a dataset, where each sentence $s$ is annotated with respect to:

\begin{itemize}
    \item an intent category $c \in \mathbb{C}$;
    \item a slot category $o_j \in \mathbb{O}$ associated with each token $t_1, ..., t_n$ of $s$, in IOB format.
\end{itemize}

\subsection{BERT}
\label{sec:bert}

Bidirectional Encoder Representations from Transformers (BERT) is an attention-based architecture to pre-train language representations.
In particular, BERT pre-trains deep bidirectional representations by jointly conditioning on both left and right contexts in a Transformer \cite{vaswani2017}. It enables transfer learning, i.e., a single architecture is pre-trained and only minimal task-specific parameters are introduced (see also \cite{peters2018,radford2018}), eliminating the need for heavily-engineered task-specific activities. The training is performed using two tasks, i.e., Masked Language Model (MLM) and Next Sentence Prediction (NSP). The former aims at capturing the properties of a language by modeling the conditional probability of $P(w_i|w_{i-1}, w_{i-2}, ..., w_1)$. Differently from a classical language model, in MLM some tokens are randomly masked to avoid a token observes itself in a multi-layered context. NSP aims at capturing useful information for sentence pair oriented tasks.

\paragraph{BERT Model.}
BERT is a \textit{Transformer Encoder} \cite{vaswani2017}, whose main building block is depicted in figure \ref{fig:transformer}. It is a multi-layered attention-based architecture, whose processing can be summarized as a sequence of \texttt{Multi-Head Attention}, \texttt{Add\&Normalization}, and \texttt{Feed-Forward} layers repeated $N$ times (with residual connections \cite{he2016}). Given a sequence of tokens $t_1, ..., t_n$, it computes a sequence of representations $h=(h_1, ..., h_n)$ to capture salient contextual information for each token. For more details, please refer to \cite{devlin2018}.

\subsection{Joint Sentence Classification and Sequence Labeling}
\label{sec:jbert}

Once BERT is pre-trained over a corpus, the learned representations model the token of a sequence in the context in which they are observed.
In order to use this model for the final tasks, e.g., classification or sequence labeling, it must be fine-tuned over a task-specific dataset. For classification tasks, Devlin and colleagues suggest using the final hidden state $h_0$ of the \cls token, which by construction should represent a fixed dimensional pooled representation of the sequence. For sequence labeling tasks, for every token $t_j$ in a sequence, the corresponding final hidden state $h_j$ can be used for classifying such token with respect to the target categories, e.g., the Named Entities.
In this work, we aim at using both the token-level and sentence-level features to perform a joint classification of the sentence and token categories.

\begin{table*}[!t]
\footnotesize
\centering
\begin{tabular}{|l|l|}
\hline
atis\_flight & showthe {[}latest{]}$_{flight\_mod}$ flight from {[}denver{]}$_{fromloc.city\_name}$ to {[}boston{]}$_{toloc.city\_name}$ \\ \hline
atis\_city & what time zone is {[}denver{]}$_{city\_name}$ in \\ \hline
atis\_flight & from {[}seattle{]}$_{fromloc.city\_name}$ to {[}salt lake city{]}$_{toloc.city\_name}$ \\ \hline
atis\_abbreviation & what does fare code {[}qx{]}$_{fare\_basis\_code}$ mean \\ \hline
\end{tabular}
\caption{Examples from the ATIS dataset. The first column indicates the intent, while the second columns contains the sentence and its slots.}
\label{tab:atisexamples}
\end{table*}

\begin{table*}[!t]
\footnotesize
\centering
\begin{tabular}{|l|l|}
\hline
SearchScreeningEvent & find [fish story]$_{movie\_name}$  \\ \hline
PlayMusic & can you play me some [eighties]$_{year}$ music by [adele]$_{artist}$ \\ \hline
AddToPlaylist & add this [track]$_{music\_item}$ to [my]$_{playlist\_owner}$ [global funk]$_{playlist}$ \\ \hline
BookRestaurant & book a spot for [3]$_{party\_size\_number}$ in [mt]$_{state}$ \\ \hline
\end{tabular}
\caption{Examples from the SNIPS dataset. The first column indicates the intent, while the second columns contains the sentence and its slots.}
\label{tab:snipsexamples}
\end{table*}

In order to achieve such goal, let us add the following parameters to the model:

\begin{itemize}
    \item $W_c \in \mathbb{R}^{|C|xH}$ and $b_c \in \mathbb{R}^{H}$ , i.e., the sentence-level classifier matrix and bias respectively;
    \item $W_o \in \mathbb{R}^{|O|xH}$ and $b_o \in \mathbb{R}^{H}$, i.e., the token-level classifier matrix and bias respectively, 
\end{itemize}

where $H$ is the dimension of the final hidden state, $|C|$ is the number of the sentence-level categories and $|O|$ is the number of token-level categories.

In order to classify a sequence $s=(t_0, ..., t_n)$ with intent $c$ and slots $o=(o_0, ..., o_n)$, each token is passed through the BERT model, resulting in a set of representations $h=(h_0, ..., h_n)$; $h_0$ is the final hidden state of the \cls token, while $h_j$ is the final hidden state of token $t_j$, for $j=1,...,n$.
The sentence-level category probabilities $P_c$ can be obtained by

$$ P_c = softmax(h_0 W^{\intercal}_c + b_c)$$

and the classified category can be obtained by $\widetilde{c}=argmax(P_c)$.

The token-level categories probabilities for the token $t_j$ can be similarly obtained through:

$$ P_o^j = softmax(h_j W^{\intercal}_o + b_o)$$

and the category for token $t_j$  can be obtained by $\widetilde{o}^j=argmax(P_o^j)$.

In standard BERT, in order to train a sentence-level classifier, we minimize the cross-entropy ($L_c$) between the predicted label $\widetilde{c}$ and the correct label $c$.
Similarly, in order to train a sequence-level classifier, for each token $t_j$ we can define the cross-entropy ($L_o^j$) between the predicted label $\widetilde{o}^j$ and the correct label $o_j$. Globally, for each sequence we can minimize the mean of the cross-entropy of each token, i.e., $L_s=\frac{1}{n}\sum_j L_o^j$.

In our setting, we aim at learning the sentence-level classifier and the token-level classifier parameters $(W_c, b_c, W_o, b_o)$ jointly. In order to fine-tune the BERT model with respect to both the tasks, we define a new loss function, which is the linear combination of $L_c$ and $L_o$ as: $$L=\alpha L_c + \beta L_o$$ where $\alpha$ and $\beta$ are new parameters to be acquired during the fine-tuning stage.
The fine-tuning will be performed through gradient descent over all the BERT model parameters plus $(W_c, b_c, W_o, b_o,\alpha,\beta)$.

\begin{table}[h]
\footnotesize
\centering
\begin{tabular}{|l|c|c|c|c|c|}
\hline
               & \textbf{Train} & \textbf{Valid} & \textbf{Test} & \textbf{Intent} & \textbf{Slot} \\ \hline
\textbf{ATIS}  & 4,478           & 500            & 893           & 26                  & 83              \\ \hline
\textbf{SNIPS} & 13,084          & 700            & 700           & 7                   & 39              \\ \hline
\end{tabular}
\caption{Datasets statistics.}
\label{tab:datastats}
\end{table}

\section{Experimental Evaluation}
\label{sec:experiments}

\begin{table*}[ht]
\footnotesize
\centering
\begin{tabular}{|l|c|c|c|c|c|c|}
\hline
 & \multicolumn{3}{c|}{\textbf{ATIS}} & \multicolumn{3}{c|}{\textbf{SNIPS}} \\ \hline
\textbf{System} & \multicolumn{1}{l|}{\textbf{Slot}} & \multicolumn{1}{l|}{\textbf{Intent}} & \multicolumn{1}{l|}{\textbf{Sentence}} & \multicolumn{1}{l|}{\textbf{Slot}} & \multicolumn{1}{l|}{\textbf{Intent}} & \multicolumn{1}{l|}{\textbf{Sentence}} \\ \hline
\texttt{Joint Seq} \cite{hakkani2016} & $.942$ & $.926$ & $.807$ & $.873$ & $.969$ & $.732$ \\ \hline
\texttt{Attention BiRNN} \cite{liu2016} & $.942$ & $.911$ & $.789$ & $.878$ & $.967$ & $.741$ \\ \hline
\texttt{IntentCapsNet} \cite{xia2018} & - & $.948$ & - & - & $.974$ & - \\ \hline
\texttt{Capsule-NLU} \cite{zhang2018} & $.952$ & $.950$ & $.834$ & $.918$ & $.977$ & $.809$ \\ \hline
\texttt{Slot-Gated FA} \cite{slotgated2018} & $.948$ & $.936$ & $.822$ & $.888$ & $.970$ & $.755$ \\ \hline
\texttt{Slot-Gated IA} \cite{slotgated2018} & $.952$ & $.941$ & $.826$ & $.883$ & $.968$ & $.746$ \\ \hline
\texttt{Slot-Gated FA}$^\dagger$ & $.951$ & $.953$ & $.844$ & $.905$ & $.970$ & $.790$ \\ \hline
\texttt{Slot-Gated IA}$^\dagger$ & $.948$ & $.951$ & $.824$ & $.881$ & $.973$ & $.769$ \\ \hline
\hline
\bertintent & - & $.971$ & - & - & $.977$ & - \\ \hline
\bertslot & $.954$ & - & - & $.959$ & - & - \\ \hline
\bertjoint & $\mathbf{.957}$ & $\mathbf{.978}$ & $\mathbf{.882}$ & $\mathbf{.962}$ & $\mathbf{.990}$ & $\mathbf{.916}$ \\ \hline
\end{tabular}
\caption{Performances over the ATIS and SNIPS datasets. Column \textit{Slot} reports the F1 of classifying the slots in a sentence. Column \textit{Intent} reports the Accuracy in finding the correct intent. Column \textit{Sentence} reports Accuracy in recognizing both the intent and all the slots. FA and IA refer to the Full and the Intent Attention variants in the Slot-Gated models. Systems marked with $\dagger$ have been re-measured in this work. All the performances are measured over the training/validation/test split as in \cite{slotgated2018}.}
\label{tab:engresults}
\end{table*}

In this section, the experimental evaluation of \bertjoint{} is discussed. First, the datasets used in the experiments is presented in section \ref{sec:dataset}. Then, in section \ref{sec:results} the experiments are discussed.

\subsection{Dataset}
\label{sec:dataset}

We conducted experiments over two benchmark datasets for the English language. As a first benchmark, we adopted the Airline Travel Information System (ATIS) \cite{atis1990}, which is a well-know benchmark for the ID and SF tasks. It contains sentences annotated with respect to intents and slots in the airline domain. In table \ref{tab:atisexamples} some examples of the sentences as well as of the annotations in the ATIS dataset are shown.

The other dataset used for evaluating the joint approach is the SNIPS dataset \cite{snips2018}. It is a collection of commands typically used in a voice assistant scenario. In table \ref{tab:snipsexamples} an excerpt of the dataset is shown.
SNIPS represents a more realistic scenario compared to the single-domain ATIS dataset. The SNIPS dataset contains more varied intents, while in the ATIS dataset all intents are from the same domain.

The two datasets represent also different training scenarios, as they differ in the number of annotated examples. The SNIPS dataset contains more than $3 \times$ number of examples with respect to ATIS. Please, see table \ref{tab:datastats} for details about the datasets.

For all the experiments in the following sections, we adopted the same dataset split as proposed by \cite{slotgated2018}.

\subsection{Experimental Setup}
\label{sec:expsetup}

In the following experiments, we adopted the multi-lingual pre-trained BERT model, which is available on the BERT authors website\footnote{\url{https://storage.googleapis.com/bert\_models/2018\_11\_23/multi\_cased\_L-12\_H-768\_A-12.zip}}. This model is composed of $12$-layer and the size of the hidden state is $768$. The multi-head self-attention is composed of $12$ heads for a total of $110$M parameters.
We adopted a dropout strategy applied to the final hidden states before the intent/slot classifiers.

We tuned the following hyper-parameters over the validation set: (i) number of epochs among ($5$, $10$, $20$, $50$); (ii) Dropout keep probability among ($0.5$, $0.7$ and $0.9$). We adopted the Adam optimizer \cite{kingma2014} with parameters $\beta_1=0.9$, $\beta_2=0.999$, L$2$ weight decay $0.01$ and learning rate $2\text{e-}5$ over batches of size $64$.

\subsection{Experimental Results}
\label{sec:results}

Table \ref{tab:engresults} reports performance measures for both ATIS and SNIPS datasets. Regarding the ID task, we computed the accuracy, while the SF performance is measured through the F1. Moreover, we report a sentence based accuracy, i.e., the percentage of sentences for which both the intent and all the slots are correct.
We compare our approach to different systems, all measured over the same training/validation/test split\footnote{\cite{liu2016} and \cite{bimodel2018} reports respectively $.982$ and $.989$ for ID and $.959$ and $.969$ for SF. However, their precise training/validation split is not known.} as reported in \cite{slotgated2018} and \cite{zhang2018}. We remeasured the performances of the two Slot-Gated systems. We adopted the available code\footnote{\url{https://github.com/MiuLab/SlotGated-SLU}} released by \cite{slotgated2018}; we tuned over the validation set the number of units among ($32$, $64$, $128$, $256$, $512$, $800$ and $1024$) with an early stop strategy over $50$ epochs. These systems are marked with the $\dagger$ symbol in the Table.
We report also the performances for two standard BERT based systems, i.e., \bertintent{} and \bertslot{}. The former is a text classifier based on the standard formulation of BERT, i.e., it learns the function $ P_c = softmax(h_0 W^{\intercal}_c + b_c)$ by only optimizing the $L_c$ loss function over the intents detection task. The latter is a sequence labeler based on the standard formulation of BERT, i.e., it learns the function $ P_o^j = softmax(h_j W^{\intercal}_o + b_o)$ by only optimizing the $L_o$ loss function over the SF task.

Regarding the ATIS dataset, notice the \bertintent{} and \bertslot{} performances with respect to the other systems. The ID performance of \bertintent{} is about $2$ points higher with respect to the best-reported system \cite{zhang2018}. The SF performance of \bertslot{} is in line, but still higher with respect to \cite{zhang2018} ($95.2$ vs $95.4$). This is in line with the findings in \cite{devlin2018}, where a very good pre-training results in an effective transfer learning in natural language tasks. This also holds with respect to the Slot-gated models \cite{slotgated2018} here re-measured. The BERT-based models allow to obtain higher performances resulting in an error reduction of about $38.0\%$ and $6.0\%$ for ID and SF, respectively.
Notice the \bertjoint{} performances on ID ($95.7$ in accuracy) and on SF ($97.8$ F1). These results confirm that the joint modeling here proposed can be beneficial also when the base is a BERT model. In fact, the performances of \bertjoint{} are higher with respect to \bertintent{} ($+$ $.3$) and \bertslot{} ($+$ $.7$) on both tasks, resulting in a straightforward $88.2$ accuracy in detecting correctly the whole sentence. This results in an error reduction of about $25\%$ for the whole sentence prediction with respect to the best-reported system.

Regarding the SNIPS dataset, we can observe very similar outcomes. Recall that the dataset size is higher and has a more varied domain with respect to the ATIS dataset. The \bertjoint{} approach set the new state-of-the-art performance over this dataset both on ID and SF, i.e., $96.2$ in accuracy and $99.0$ in F1 respectively. As a consequence, also the overall sentence accuracy set the new state-of-the-art, i.e., $91.6$ accuracy in detecting correctly the intents and all the slots for a sentence (error reduction of about $30\%$ with respect to the best-reported system).
Again, \bertintent{} and \bertslot{} approaches are performing very well on this task, but, again, the joint model here proposed is beneficial.

\subsection{Measuring the Impact of Joint Modeling}
\label{sec:learningcurves}

\begin{figure}[!ht]
  \centering
    \includegraphics[width=0.48\textwidth]{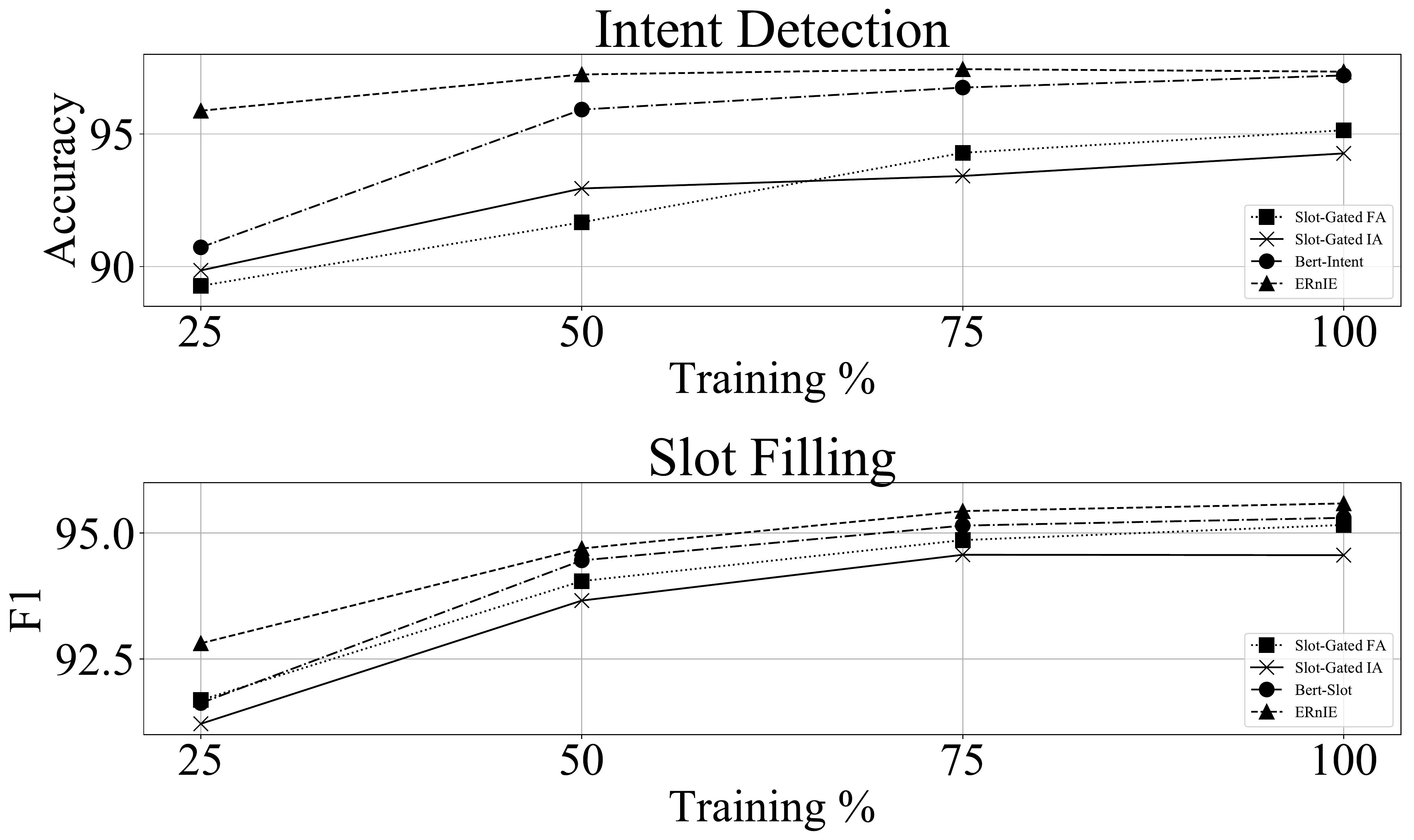}
  \caption{Learning curves for the ATIS dataset.}
  \label{fig:atislearning}
\end{figure}

As discussed in section \ref{sec:results}, \bertjoint{} is very effective in classifying intents and slots. In fact, intents and slots are strongly related. In order to better understand what is the contribution of a joint approach, in this section we provide the analysis of how fast a joint model reaches higher performances with respect to non-joint approaches. We, thus, compare the performances of the BERT-based systems \bertintent, \bertslot{} with respect to \bertjoint{} on poor training conditions.
That is, we trained each of this model on training sets of growing sizes. In particular, we trained the models on $25\%$, $50\%$, $75\%$ and $100\%$ of the training data for both ATIS and SNIPS datasets.
We performed this evaluation with the best hyper-parameters found for the evaluations in section \ref{sec:results}. We acquired the models over $5$ different shuffles of the training set and we report the averaged results.
Moreover, we report the same evaluations with the two Slot-Gated systems.

\begin{figure}[!ht]
  \centering
    \includegraphics[width=0.48\textwidth]{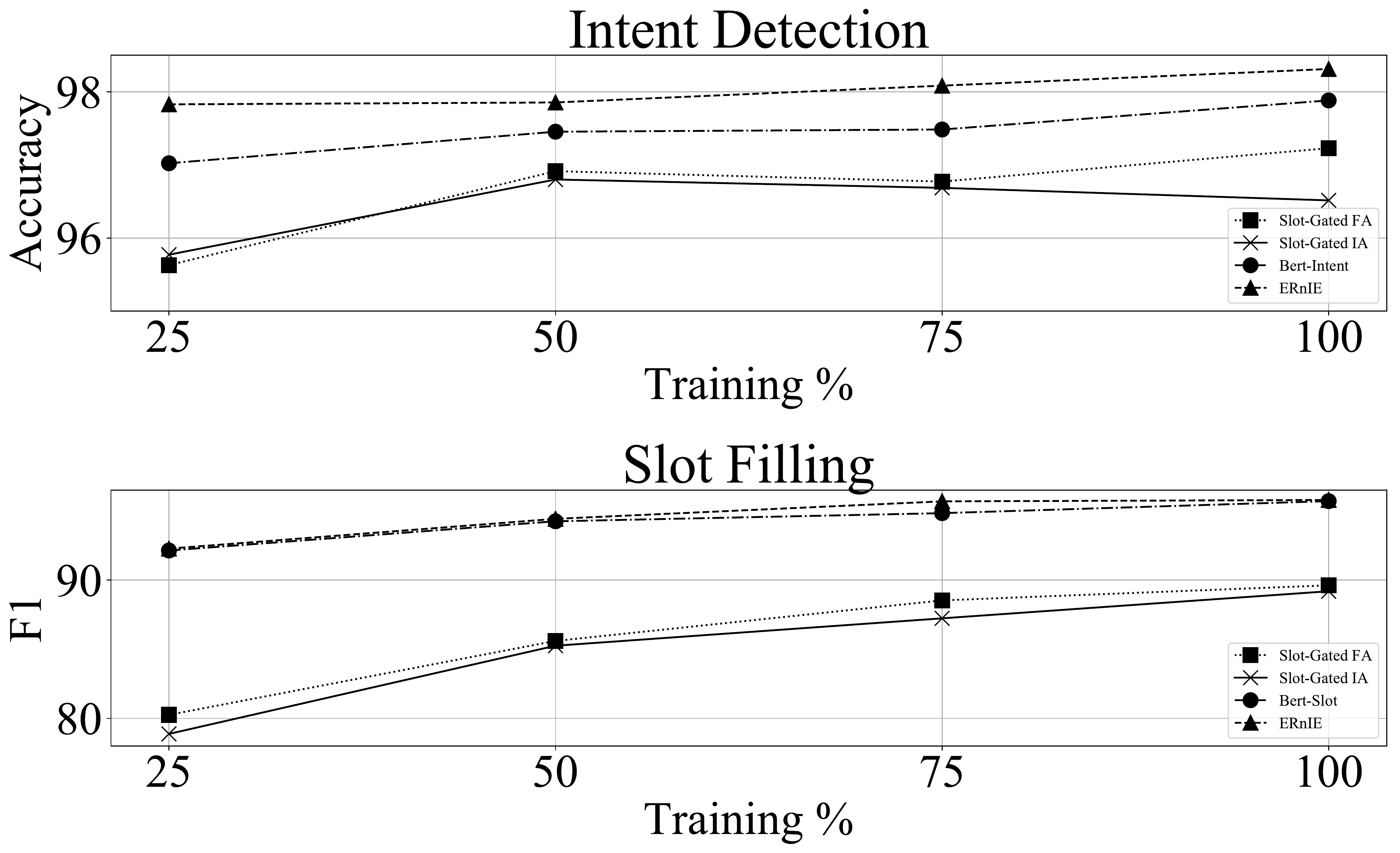}
  \caption{Learning curves for the SNIPS dataset.}
  \label{fig:snipslearning}
\end{figure}

Figure \ref{fig:atislearning} shows learning curves for the ATIS dataset. First, notice that the BERT-based systems are performing better than the Slot-gated models at all training set sizes. This confirms that a good pre-training is beneficial in any training condition.
Moreover, notice that the benefits of using a joint approach: starting with only $25\%$ of the training material \bertjoint{} is beneficial both for the ID and SF. In ID, at $25\%$ of the training set, there is a difference of about $5$ points ($90.7$ vs $95.8$) between \bertintent{} and \bertjoint{} (about $54\%$ relative error reduction). A similar outcome can be observed for \bertslot{} vs. \bertjoint{}, where the difference is about of $1.2$ points in the F1 measure (about $14\%$ relative error reduction).
When the training set size grows, the performances of the systems are more similar, but a clear advantage of the \bertjoint{} can be always observed.

Figure \ref{fig:snipslearning} shows learning curves for the SNIPS dataset. Again, there is a benefit in using a joint approach at lower training set sizes both for ID and SF. Again, it seems that the ID task benefits more of the joint modeling. At $25\%$ of annotated material (i.e., about $3,271$ example), the \bertjoint{} model outperforms the \bertintent{} model of about $1$ point in accuracy ($97.0$ vs $97.8$), resulting in about $26\%$ relative error reduction. Instead, the \bertjoint{} outperforms \bertslot{} of about $0.2$ points in F1 with $25\%$ of the training material, resulting in about $2.5\%$ relative error reduction. Even with the SNIPS dataset we can observe that when the training material grows, all the models perform better, but with a clear advantage of our joint approach.


\section{Detecting Intent and Slots in Italian}
\label{sec:expitalian}

In order to verify whether \bertjoint{} can be applied to a different language, we evaluate it on an Italian dataset. We aim at checking whether the multilingual capability of BERT is preserved also when facing a joint learning task. In the following, we provide the dataset description in section \ref{sec:itadataset}; then, we discuss the experiments in section \ref{sec:itaexp}.

\subsection{Dataset}
\label{sec:itadataset}

To  the  best  of  our  knowledge,  there  is  no  annotated dataset for SLU in Italian. 
In order to obtain a good quality resource we derived it from an existing one of another language. We used the SNIPS dataset as a starting point for these reasons: i) it contains a reasonable amount of examples; ii) it is multi-domain; iii) we  believe it could represent a more realistic setting in today's voice assistants scenario. We performed a semi-automatic process consisting of two phases: an automatic translation of the sentences with contextual alignment of intents and slots; a manual validation of the translations and annotations.
In the first phase, we translated each English sentence in Italian by using the Translator Text API, which is part of the Microsoft Azure Cognitive Services \footnote{\url{https://docs.microsoft.com/en-us/azure/cognitive-services/translator/translator-info-overview}}. The intent associated with the  English sentence has been copied to its Italian counterpart. Slots have been transferred by using the alignment of source and target tokens provided by the Translator Text API.
In order to create a more valuable resource in Italian, we also performed an automatic substitution of the name of movies, movie theatres, books, restaurants and of the locations with some Italian counterpart. First, we  collected a set $E$ from the Web about $20,000$ Italian version of such entities; then, we substituted each entity in the sentences of the dataset with one randomly chosen from $E$.
In the second phase, the dataset was split into different sets, and each has been annotated by one annotator and reviewed by another annotator. A further review was performed in case of disagreement between the annotators.
Some interesting phenomena emerge for the different intents. The translation of \textit{GetWeather}'s sentences was problematic because the main verb is often misinterpreted, while in the sentences related to the intent \textit{BookRestaurant} a frequent failure occurred on the interpretation of prepositions. For example, the sentence “Will it get chilly in North Creek Forest?” is translated as “Otterrà freddo in North Creek Forest?”: while the correct translation is “Sarà freddo a North CreekForest?”. The verb ``get'' in Italian can be translated in different ways depending on the context. In this case, the system misinterpreted the context, assigning to ``get'' the wrong meaning.

Finally, with this approach we obtained an Italian dataset (\snipsita) composed of 7,142 sentences annotated with respect to intents and slots, almost equally distributed on the different intents. The effort spent on the construction of this new resource, according to the procedure described, is about 24 FTE\footnote{Full Time Equivalent}, with an average production of about 300 sentences per day. We consider this effort lower than typical efforts to create linguistic resources from scratch.

\subsection{Experimental Results}
\label{sec:itaexp}

We selected from \snipsita{} the same train/validation/test split used for the English evaluations. It results in $5,742$, $700$ and $700$ respectively for training, validation and test. We run the experiments with the same setup used in the English scenario: we tuned the number of epochs ($10$, $20$ and $50$) and the dropout parameter ($0.5$, $0.7$ and $0.9$), and we used the same settings for the Adam optimizer. We compare \bertjoint{} to the non-joint versions of BERT, i.e., \bertintent{} and \bertslot. Moreover, we compare also with the Slot-Gated models. We adopted the available code released by the authors; we tuned for these models the number of units among ($32$, $64$, $128$, $256$, $512$, $800$ and $1024$) with early stop over $50$ epochs.

\begin{table}[!ht]
\footnotesize
\centering
\begin{tabular}{|l|c|c|c|}
\hline
\textbf{System} & \multicolumn{1}{l|}{\textbf{Slot}} & \multicolumn{1}{l|}{\textbf{Intent}} & \multicolumn{1}{l|}{\textbf{Sentence}} \\ \hline
\texttt{Slot-Gated FA} & $.810$ & $.959$ & $.611$ \\ \hline
\texttt{Slot-Gated IA} & $.809$ & $.963$ & $.613$ \\ \hline
\hline
\bertintent & - & $.967$ & - \\ \hline
\bertslot & $.897$ & - & - \\ \hline
\bertjoint & $\mathbf{.900}$ & $\mathbf{.976}$ & $\mathbf{.771}$ \\ \hline
\end{tabular}
\caption{Performances over \snipsita{}. Column \textit{Slot} reports the F1 of classifying the slots. Column \textit{Intent} reports the Accuracy in finding the intent. Column \textit{Sentence} reports Accuracy in recognizing both the intent and all the slots. FA and IA refer to the Full and the Intent Attention variants in the Slot-Gated models.}
\label{tab:itaresults}
\end{table}

\begin{table*}[ht]
\footnotesize
\centering
\begin{tabular}{|l|c|c|c|c|c|c|}
\hline
 & \multicolumn{3}{c|}{\textbf{English}} & \multicolumn{3}{c|}{\textbf{Italian}} \\ \hline
\textbf{System} & \multicolumn{1}{c|}{\textbf{Slot}} & \multicolumn{1}{c|}{\textbf{Intent}} & \multicolumn{1}{c|}{\textbf{Sentence}} & \multicolumn{1}{c|}{\textbf{Slot}} & \multicolumn{1}{c|}{\textbf{Intent}} & \multicolumn{1}{c|}{\textbf{Sentence}} \\ \hline
\bertintent & - & $\mathbf{.981}$ & - & - & $\mathbf{.980}$ & - \\ \hline
\bertslot & $.952$ & - & - & $\mathbf{.904}$ & - & - \\ \hline
\bertjoint & $.961$ & $.980$ & $.910$ & $\mathbf{.913}$ & $\mathbf{.983}$ & $\mathbf{.794}$ \\ \hline
\end{tabular}
\caption{Multi-lingual experiments: each system is trained over both English and Italian training sets and tested separately over English and Italian. Column \textit{Slot} reports the F1 of classifying the slots in a sentence. Column \textit{Intent} reports the Accuracy in finding the correct intent. Column \textit{Sentence} reports Accuracy in recognizing both the intent and all the slots.}
\label{tab:multiresults}
\end{table*}

In table \ref{tab:itaresults} the performances of the systems are shown. The slot performance is the F1 while the Intent and Sentence performances are measured with the accuracy. Notice that all models are performing similarly to their English counterpart\footnote{The training set size is about $50\%$ of the English dataset, thus the Italian measures must be compared with the English measures at about $50\%$ of the learning curve.}. First, notice the performances of the Slot-gated models \cite{slotgated2018} over this dataset. Regarding the SF task, the new language seems to be critical for both the variants of the model, as they reach only about $81.0\%$ in F1. Notice that in similar settings, i.e., about $50\%$ of the training examples, the English performance was about $5$ points higher. Regarding the ID task, the performances are instead higher also for this language.
Again, we can observe that the \bertjoint{} training is beneficial for obtaining higher ID performances with respect to the model without the joint modeling (i.e., \bertintent{} and \bertslot{}). Also the SF task benefits from the adoption of joint training. Notice that, the proposed approach outperforms the Slot-Gated models. This is a straightforward result as no modification to the model has been made for the Italian language.

\subsection{Multi-lingual Detection of Intent and Slots}
\label{sec:multilingual}

As pointed out in section \ref{sec:dataset} the \snipsita{} dataset is obtained with a low-effort process. This results in performances that are lower with respect to their English counterpart\footnote{There is a difference of about $8$ points in correctly determining a whole sentence accuracy (intent+slot)}.
One could think of exploiting the BERT multi-lingual capabilities to train an SLU system on the English language and to use it to generate annotations in Italian in order to obtain a higher quality dataset or to increase the size of annotated examples. However, such a system would fail\footnote{We performed a cross-lingual experiment by training on one language and testing over the other. The performances of ID could be considered satisfactory (about $89\%$ by training in English and testing in Italian and about $79\%$ vice-versa. However, the slot recognition is far worse, i.e., about $60\%$ and $55\%$, respectively.} in correctly recognizing the slots. In fact, they are very different in the two languages as both their lexical surface and the syntax is highly language-specific.

For these reasons, we believe that a more consistent way of exploiting the capabilities of the BERT model is to train a multi-lingual model over both the datasets. In this way, we aim at injecting into a low-effort dataset (\snipsita) the information contained in a higher quality dataset (SNIPS). In Table \ref{tab:multiresults} we provide the experimental results of such a setting. We trained the \bertintent, \bertslot{} and \bertjoint{} models with both the English and Italian training sets and we tested over the two test sets separately.
Notice that the performances over English are slightly worse but comparable with the monolingual training (see Table \ref{tab:engresults}). It demonstrates that the multi-lingual setting doesn't degrade too much the performances on that language. However, notice that the performances over the Italian dataset are higher, resulting in about $2$ points gain. Notice how the accuracy in predicting correctly a whole sentence increases from $77.1$ to $79.4$, which is about a $10\%$ error reduction. Again, our joint approach performances are higher with respect to the non-joint versions of the model.
Moreover, a multi-lingual SLU model is also more efficient for production purposes. In fact, there will be the need for deploying only one model for multiple languages, resulting in architectural savings.

\section{Related Work}
\label{sec:related}

\paragraph{Intent Detection.}
The ID task is addressed as a text classification problem, in which classical machine learning or deep learning have been widely adopted. Many researchers employed support vector machines \cite{chelba2003}, or boosting-based classifiers \cite{schapire2000}.
Recently, many works exploited Deep Learning ability to learn effective representations. For example \citeauthor{sarikaya2011} uses Deep Belief Nets (DBNs) for natural language call routing, where a multi-layer generative model is learned from unlabeled data. Then, the features discovered are used to pre-initialize a feed-forward network which is fine-tuned on labeled data. In \cite{xia2018} ID is addressed in a Zero-shot \cite{xian2017} framework with Capsule Networks \cite{hinton2011}.

\paragraph{Slot Filling.}
The SF task is addressed through supervised sequence labeling approaches, e.g., MEMMs \cite{mccallum2000}, CRF \cite{raymond2007} or, again, with Deep Learning, such as Recurrent Neural Networks (RNNs) \cite{hochreiter1997long}.
Deep learning research started as extensions of Deep Neural Networks and DBNs (e.g., \cite{deoras2013}) and is sometimes merged with Conditional Random Fields \cite{xu2013}.
Later, \citeauthor{mesnil2015} proposes models based on recurrent neural networks (RNNs). On the same line of research is the work of \cite{liu2015}, which, uses RNNs but introduces label dependencies by feeding previous output labels.
\citeauthor{chen2016} address the error propagation problem in a multi-turn scenario by means of an End-to-End Memory Network \cite{sukhbaatar2015} specifically designed to model the knowledge carryover.

\paragraph{Joint Models.}
Recently, ID and SF have been addressed by jointly modeling the two into a unique architecture \cite{hakkani2016,liu2016,bimodel2018,slotgated2018,zhang2018}. In fact, it has been found that a model that is trained on both tasks jointly, can achieve better performances on both. For example, in \cite{hakkani2016} a single RNN architecture for domain detection, ID and SF in a single SLU model is proposed showing gains in each.
In \cite{liu2016}, ID and SF are investigated through an attention-based \cite{bahdanau2014} mechanism within an encoder-decoder framework. In \cite{bimodel2018}, a Bi-model based RNN combines two task-specific networks, i.e., a Bidirectional LSTM and an LSTM decoder; they are trained without a joint loss function. 
\citeauthor{slotgated2018} extend an attention-based model for the joint task of ID and SF. In particular, a slot gate focuses on learning the relationship between intent and slot attention vectors. In \cite{zhang2018} Capsule Networks \cite{hinton2011} are adopted to jointly classify ID and SF through a hierarchical capsule network structure. This should capture the inter-dependencies between words/slot and intents in a hierarchy of feature detectors.

\section{Conclusion}
\label{sec:conclusion}

In this work, we addressed the problem of Intent Detection and Slot Filling in Spoken Language Understanding. We based on the BERT model ability to provide effective pre-trained representations. We adapted the original BERT fine-tuning to define a new joint learning framework.
\bertjoint{} acquires very effective representations for a joint learning task. We provided an extensive evaluation in the English language: BERT-based approaches performs very well on the intent detection and slot filling tasks. \bertjoint{} learning schema provides even better results, i.e., the new state-of-the-art for these tasks.
Moreover, we showed that this approach is beneficial when less annotated data is available. We also showed the multi-lingual capability of the model for dealing with the Italian language. We annotated a new SLU dataset in Italian, and we measured over it the performances of our approach: in this setting and in the multi-lingual setting, \bertjoint{} outperforms non-joint approaches.
In future, we aim at investigating languages with very different structures, e.g., Chinese or Arabic. It could also be interesting to adapt our model to multi-intent scenarios, or to model other semantic phenomena, e.g., jointly classifying frames and semantic arguments in Frame Semantics \cite{fillmore85}.

\bibliography{acl2019}
\bibliographystyle{acl_natbib}

\end{document}